\documentclass[letterpaper, 10 pt, conference]{ieeeconf}  

\IEEEoverridecommandlockouts                              

\usepackage{graphics} 
\usepackage{epsfig} 
\usepackage{times} 
\usepackage{amsmath} 
\usepackage{amssymb}  
\usepackage{cite}
\usepackage{multirow}
\usepackage{booktabs}
\usepackage[table]{xcolor}
\usepackage{xurl}
\usepackage{hyperref}

\definecolor{LightRed}{RGB}{255,230,230}
\definecolor{LightBlue}{RGB}{225,238,250}
\definecolor{LightGreen}{RGB}{226,239,218}

\newcommand{\SEthree}{\mathrm{SE}(3)}

\title{\LARGE \bf
DualManip: Agentic Dynamic Manipulation via Dual-Path
Semantic Reasoning and Geometric Adaptation
}

\author{Chengxi Li$^{1,*}$, Yan Di$^{2,*}$, Yingyue Li$^{1,*}$, Ruida Zhang$^{1}$, Mingyang Li$^{3}$, Xiangyang Ji$^{1,\dagger}$%
\thanks{$^{1}$ Chengxi Li, Yingyue Li, Ruida Zhang and Xiangyang Ji are with the Department of Automation, Tsinghua University, China. {\tt\small \{lichengx21, yingyue-21, zhangrd23\}@mails.tsinghua.edu.cn, xyji@tsinghua.edu.cn}}
\thanks{$^{2}$ Yan Di is with the School of Computer Science and Technology, Harbin Institute of Technology (Shenzhen), China. {\tt\small diyan@hit.edu.cn}}
\thanks{$^{3}$ Mingyang Li is with the Beijing Institute of Control Engineering, China. {\tt\small lmy\_hit@163.com}}
\thanks{
$^{\dagger}$ Corresponding author}
\thanks{
$*$ Equal contribution}
}%

\begin{document}

\maketitle
\thispagestyle{empty}
\pagestyle{empty}


\begin{abstract}
Vision-language models (VLMs) enable open-vocabulary reasoning for robot manipulation, but their high inference latency limits responsiveness in dynamic scenes.
Many scene changes, however, alter object geometry without invalidating task intent.
We present DualManip, a dual-path framework that decouples infrequent semantic reasoning from responsive geometric adaptation. 
The semantic path decomposes the task and grounds task-relevant interactions, followed by a constraint-solving module for pose optimization.
During execution, the geometric path continuously updates template-to-observation correspondences from live RGB-D observations via a shape-adaptive network.
These correspondences transfer task-relevant grasp contacts across observations, enabling online grasp reconstruction under object motion and non-rigid deformation.
The Information Interaction Module bridges the two paths by initializing task-relevant grasps from semantic grounding, validating geometric updates, and triggering semantic replanning upon update failures.
Real-world evaluation spans six manipulation tasks covering non-rigid deformation, articulated reconfiguration, rigid motion, and high-precision assembly across three settings: static, single-change, and continuous dynamic.
DualManip demonstrates superior manipulation robustness, particularly under continuous scene changes, while achieving geometric adaptation approximately 46$\times$ faster than agentic verification and semantic replanning.
Our project page: \url{https://lichengxi1.github.io/Dualmanip}.

\end{abstract}

\section{INTRODUCTION}
Robots operating in real-world environments must manipulate objects while the scene continues to evolve during perception, planning, and execution~\cite{domino,dynamicvla,dynamicwam}.
Objects may be displaced, remain in motion, or undergo substantial deformation, especially in household and human-centered settings involving garments, ropes, soft objects, and articulated structures~\cite{demavla,unigarmentmanip,sparsemeetsdense,vitacman}.
Compared with static rigid-object manipulation, these scenarios require both rapid perception--action updates and the ability to accommodate time-varying, non-rigid geometry.
Responsive manipulation of dynamic and deformable objects therefore remains an important challenge for general-purpose robotic systems.

End-to-end Vision-Language-Action (VLA) models and World-Action Models (WAMs) offer a promising paradigm for dynamic manipulation by learning closed-loop policies directly from large-scale robot data~\cite{dynamicvla,demavla,dswam,dynamicwam}.
However, these approaches typically require considerable training data tailored to specific tasks or robot embodiments and still incur the inference cost of large models.
An alternative is a modular paradigm, in which pretrained Vision-Language Models (VLMs) provide open-vocabulary task reasoning while explicit geometric modules perform spatial grounding, motion generation, and low-level control~\cite{voxposer,copa,geomanip,seam,tiptop,unimanip,rekep,omnimanip}.
By separating high-level reasoning from geometric execution, these systems provide interpretable interfaces and often generalize effectively to novel objects and instructions.
Their ability to adapt to dynamic scenes, however, remains limited.
Existing methods typically handle scene changes through geometric tracking, state verification, or episodic replanning~\cite{rekep,omnimanip,closedloopgrasp,unimanip,clea}.
Such mechanisms struggle in continuously evolving scenes: geometric methods (e.g., rigid pose or sparse point tracking) may degrade under non-rigid deformation, while agentic verification and replanning incur excessive latency.


\begin{figure}[t]
    \centering
    \includegraphics[width=0.85\columnwidth]{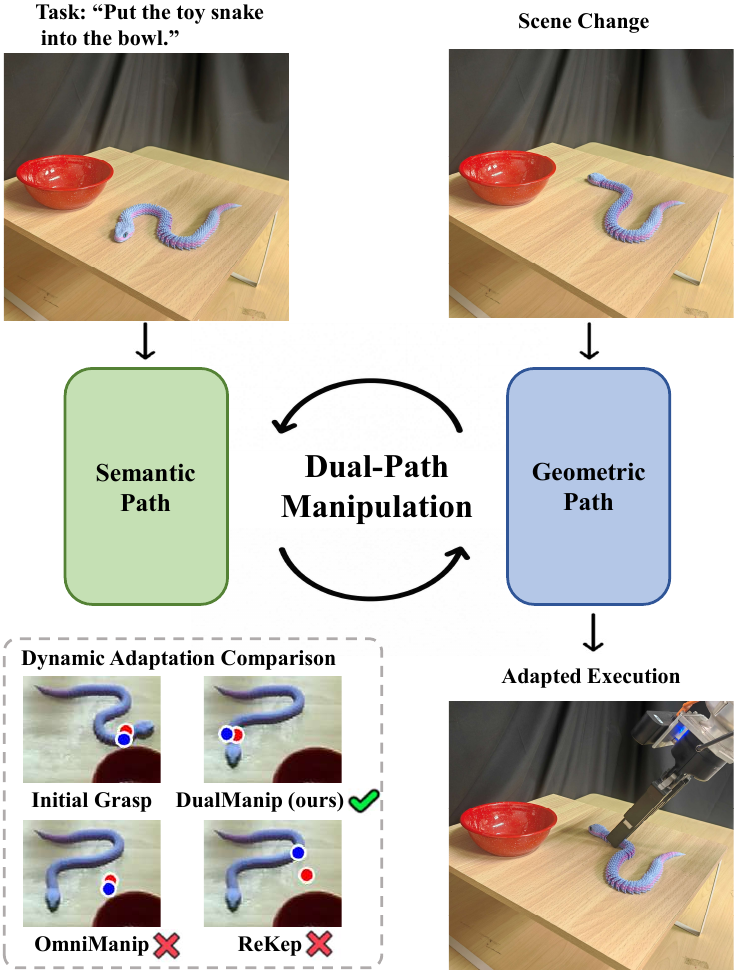}
    \caption{
        DualManip decouples semantic reasoning from geometric grasp adaptation. As the toy snake deforms, the geometric path updates the grasp while preserving the task intent established by the semantic path. In contrast, ReKep and OmniManip fail to maintain a valid task-consistent grasp under the same non-rigid deformation.
    }
    \label{fig:teaser}
    \vspace{-4mm}
\end{figure}

The key observation is that scene geometry can change without invalidating task intent, as illustrated in Fig.~\ref{fig:teaser}.
We distinguish \emph{geometric changes}, where task intent remains valid but prior scene geometry becomes outdated, from \emph{semantic changes}, which alter task interpretation and require re-reasoning.
Existing agentic systems lack an efficient mechanism to distinguish between them, causing frequent geometric changes to either trigger unnecessary high-latency VLM inference or leave robots operating on stale geometry.

To this end, we present DualManip, a \emph{dual-path} architecture that separates deliberative semantic reasoning from responsive geometric adaptation.
The \emph{semantic path} performs VLM-based reasoning over the instruction and scene observation, producing a task-conditioned manipulation plan and grounding task-relevant object regions.
The constraint-solving module then generates action constraints from these semantic outputs and optimizes for an executable pose.
During execution, the \emph{geometric path} continuously updates the scene geometry from live RGB-D observations and adapts the planned grasp without repeating semantic reasoning.
Specifically, DualManip transfers task-relevant contact points from a pre-scanned object template to the current observation through point correspondence, and reconstructs the grasp pose from the transferred contacts under object motion and non-rigid deformation.
The adapted grasp is then verified for reachability and collision safety before execution, while unreliable correspondence or infeasible geometry triggers a fallback to renewed semantic grounding and planning.
By decoupling task semantics from time-varying geometry, DualManip preserves manipulation intent established by the VLM while enabling low-latency, online adaptation to continuously evolving scenes, including non-rigid deformation.

We evaluate DualManip on six real-world manipulation tasks spanning non-rigid deformation, articulated reconfiguration, rigid motion, and precision assembly.
We further consider three evaluation settings: \emph{static}, where the scene remains unchanged; \emph{single-change}, where the object undergoes a single motion or deformation before remaining stationary; and \emph{continuous dynamic}, where the object continues to move or deform during execution.
Experimental results across diverse tasks confirm DualManip's superior manipulation robustness, particularly under scene changes, while maintaining markedly lower adaptation latency than an agentic verification-and-replanning baseline.

The main contributions are summarized as follows:
\begin{itemize}
    \item We propose DualManip, a dual-path framework that combines agentic semantic planning with responsive geometric adaptation, enabling intent-consistent action updates under dynamic scene variations.

    \item We develop a shape-adaptive correspondence network with geometry-aware learning that jointly models rigid motion and non-rigid deformation, enabling robust point-level matching across different object states.

    \item We introduce a task-conditioned contact-transfer and feasibility-routing mechanism for accurate grasp reconstruction, reliable feasibility validation, and selective fallback to semantic replanning.
\end{itemize}

\section{RELATED WORK}

\subsection{Foundation Models for Robot Manipulation}
End-to-end VLA policies directly map vision-language observations to robot actions, enabling increasingly general closed-loop manipulation~\cite{rt2,openvla,pi05,dynamicvla,demavla}.
WAMs further incorporate predictive world representations for long-horizon and dynamic execution~\cite{tau_0,oawam,dswam,dynamicwam,Abot}.
Despite their strong capabilities, these approaches typically rely on considerable task- or embodiment-specific robot data and incur high training and inference costs.
A complementary paradigm uses pretrained VLMs~\cite{gpt5-6, qwen3.5, llava} for semantic reasoning together with explicit geometric or planning representations, including 3D value maps~\cite{voxposer}, part-level spatial constraints~\cite{copa,geomanip}, relational keypoint constraints~\cite{rekep,moka}, object-centric primitives~\cite{omnimanip}, semantic-to-action representations~\cite{seam,grace}, and task-and-motion planning~\cite{tiptop}.
DualManip follows this modular paradigm, targeting dynamic manipulation where object motion and non-rigid deformation require continuous geometric adaptation.

\subsection{Agentic and Dynamic Manipulation}
Recent agentic manipulation systems improve closed-loop execution through observation, verification, memory, tool orchestration, and recovery~\cite{clea,agentchord,roboreact,goal2skill,unimanip,vlasastools,harnessvla}.
CLEA~\cite{clea} and UniManip~\cite{unimanip} maintain task and scene states for adaptive replanning, while AgentChord~\cite{agentchord} and RoboReact~\cite{roboreact} respond to execution disturbances through preplanned recovery or object-centric skill re-grounding.
Other systems combine VLM- or LLM-based orchestration with VLA policies for low-level execution~\cite{goal2skill,vlasastools,harnessvla}.
Despite improved closed-loop reasoning, adaptation typically occurs at the task or skill level and relies on discrete replanning.
DualManip instead targets execution-time geometric changes, continuously adapting the planned grasp through a dedicated geometric path while preserving valid task semantics.

\subsection{Geometric Tracking and Deformable Correspondence}
Rigid object tracking models temporal motion with a single $\SEthree$ transform~\cite{se3tracknet,foundationpose}, while point tracking propagates correspondences in 2D or 3D space~\cite{pips,cotracker3,spatialtracker}.
For deformable  geometry, correspondence can be established through canonical deformation~\cite{3dcoded,cadex}, learned point matching~\cite{lepard,dvmatcher}, or dense optical flow~\cite{raft,gmflow}.
These representations have supported rigid pose updates~\cite{omnimanip,closedloopgrasp}, task-relevant point tracking~\cite{rekep}, and deformable manipulation~\cite{unigarmentmanip,sparsemeetsdense}.
In deformable manipulation, however, existing methods are often tailored to specific object categories or interaction patterns.
Despite an object-level template prior, DualManip employs a unified template-to-observation correspondence mechanism for execution-time action transfer.

\section{METHOD}

\begin{figure*}[t]
    \centering
    \includegraphics[width=\textwidth]{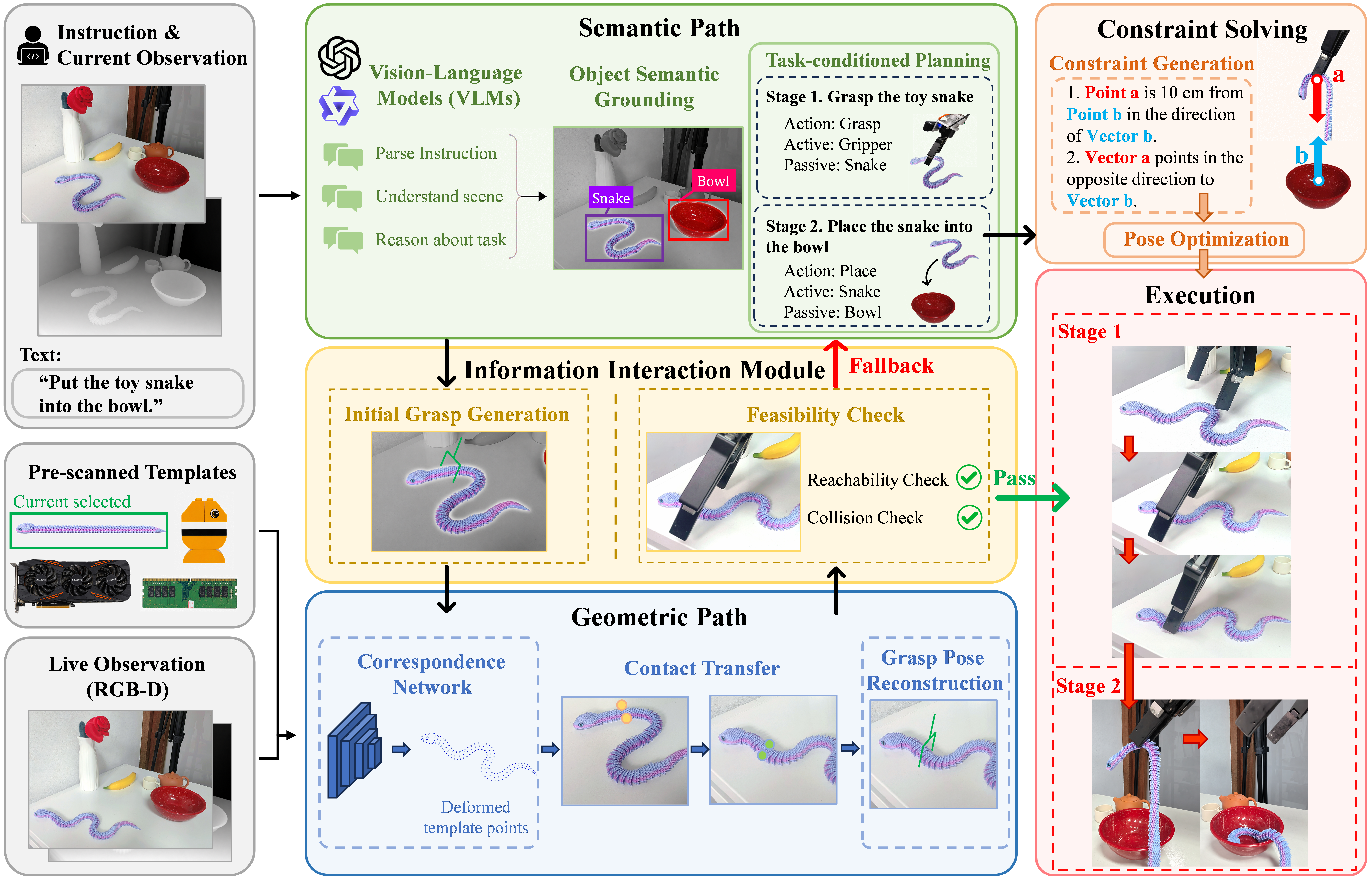}
    \caption{
        Overview of DualManip.
        Given instruction and RGB-D observation, the semantic path establishes a task-conditioned plan, from which the information interaction module generates the initial grasp.
        During execution, the geometric path continuously updates the grasp from live RGB-D observations, executing feasible updates and falling back otherwise.
        For subsequent manipulation stages, task-conditioned spatial relations are formulated as constraints and optimized into executable poses.
    }
    \label{fig:overview}
    \vspace{-4mm}
\end{figure*}

\subsection{Overview}

Many scene changes alter object geometry without changing task intent. DualManip therefore separates semantic reasoning from geometric adaptation, as shown in Fig.~\ref{fig:overview}. The semantic path produces a staged plan and grounds task-relevant objects and regions. The geometric path transfers grasp contacts to the current geometry through template-to-observation correspondence, without repeating semantic reasoning. An information interaction module initializes the grasp, validates geometric updates, and invokes semantic replanning when adaptation is unreliable or infeasible. For subsequent manipulation stages, constraint solving converts task-conditioned spatial relations into executable poses.
Together, these components allow DualManip to preserve valid task semantics while responding efficiently to object motion and non-rigid deformation.


\subsection{Semantic Path and Constraint Solving}
Given a manipulation instruction, the \emph{Semantic Path} decomposes the task into multiple stages and identifies the task-relevant objects.
For each stage, it grounds interaction locations and directions, while the constraint-solving module formulates spatial constraints, and optimizes for an executable end-effector pose.

\textbf{Task Parsing.}
The manipulation task is first decomposed by a VLM~\cite{gpt5-6,qwen3.5}, yielding a sequence of stages $\mathcal{S}=\{\mathcal{S}_i\}_{i=1}^{n}$.
Following OmniManip~\cite{omnimanip}, each stage is represented as $\mathcal{S}_i=\{a_i,\mathcal{O}_i^{a},\mathcal{O}_i^{p}\}$,
where $a_i$ denotes the action, and $\mathcal{O}_i^{a}$ and $\mathcal{O}_i^{p}$ denote the active and passive objects, respectively.
We then use Segment Anything Model 3 (SAM 3)~\cite{sam3} to localize these objects and produce their segmentation masks.
The corresponding object point clouds $\mathbf{P}_i^{a}$ and $\mathbf{P}_i^{p}$ are recovered from the masks and depth observation.

\textbf{Interaction Grounding.}
We characterize each interaction by a task-relevant location and direction.
For stage $\mathcal{S}_i$, the VLM identifies functional regions on both active and passive objects.
Their descriptions are used to prompt SAM 3 on object-centric image crops, yielding local masks and the corresponding 3D interaction points $\mathbf{p}_i^{a}$ and $\mathbf{p}_i^{p}$.
To obtain interaction directions, we employ Sofar~\cite{sofar} to infer task-conditioned semantic orientations $\mathbf{v}_i^{o}$ from the object point cloud $\mathbf{P}_i^{o}$ and region description $l_i^{o}$:
\begin{equation}
\mathbf{v}_i^{o}
=
\operatorname{Sofar}(\mathbf{P}_i^{o}, l_i^{o}),
\quad o\in\{a,p\}.
\end{equation}

\textbf{Constraint Generation.}
Utilizing these grounded interaction locations and directions, we formulate the desired interaction into explicit spatial constraints.
Inspired by CoPa~\cite{copa}, we predefine a compact set of common geometric relations, encompassing point distances and directional alignments (e.g., perpendicularity and collinearity). 
We overlay the grounded points and directional vectors onto the scene image and query the VLM with the annotated image, stage intent, and predefined relation set to obtain the stage-specific constraint set 
\begin{equation}
\mathcal{C}_i=\{C_{i,k}\}_{k=1}^{m_i},
\end{equation}
where $m_i$ is the number of constraints for stage $i$.

\textbf{Pose Optimization.}
Each constraint $C_{i,k}$ is converted into a geometric violation cost $E_{i,k}(\mathbf{T})$ over the end-effector pose $\mathbf{T}\in\SEthree$.
The target pose is obtained by minimizing the total constraint violation:
\begin{equation}
\mathbf{T}_i^{*}
=
\arg\min_{\mathbf{T}\in\SEthree}
\sum_{k=1}^{m_i}
\lambda_{i,k}E_{i,k}(\mathbf{T}),
\end{equation}
where $\lambda_{i,k}$ controls the contribution of each constraint.

\subsection{Geometric Path for Dynamic Adaptation}
\label{sec:geometric_adaptation}

The \emph{Geometric Path} enables low-latency grasp adaptation by maintaining point-level correspondences between a pre-scanned object template and live RGB-D observations, through which task-relevant grasp contacts are transferred to the current geometry. 
Unlike rigid pose or sparse point tracking, this template-based correspondence provides more robust handling of both object motion and non-rigid deformation.

\textbf{Template Construction.}
For each manipulated object, we capture multi-view images and reconstruct a 3D template via Hunyuan3D~\cite{hunyuan3d}, which is sampled into a reference point cloud $\mathbf{P}^{\mathrm{ref}}\in\mathbb{R}^{M\times3}$ as a persistent geometric reference.
At time $t$, the current object point cloud $\mathbf{P}_t\in\mathbb{R}^{N_t\times3}$ is extracted from the latest RGB-D frame, where $M$ and $N_t$ denote the numbers of template and observed points, respectively.
Both $\mathbf{P}^{\mathrm{ref}}$ and $\mathbf{P}_t$ are jointly processed by a correspondence network for point-level matching.

\textbf{Shape-Adaptive Point Correspondence.}
To overcome severe geometric discrepancies caused by object motion and non-rigid deformation, inspired by~\cite{spd,sgpa}, we propose a shape-adaptive correspondence network that dynamically aligns dense points across disparate object states.
Given the reference point cloud $\mathbf{P}^{\mathrm{ref}}$ and the current observation $\mathbf{P}_t$, let
$\mathbf{F}^{\mathrm{ref}}_t\in\mathbb{R}^{M\times d}$ and
$\mathbf{F}_t\in\mathbb{R}^{N_t\times d}$ denote their learned point-level features for correspondence estimation, where $d$ is the feature dimension.
After point-wise $\ell_2$ normalization, their pairwise similarities are normalized through Sinkhorn iterations~\cite{sinkhorn} to obtain a soft assignment matrix
\begin{equation}
\mathbf{A}_t
=
\operatorname{Sinkhorn}
\left(
\frac{\mathbf{F}_t(\mathbf{F}^{\mathrm{ref}}_t)^\top}{\tau}
\right)
\in\mathbb{R}^{N_t\times M},
\end{equation}
where $\tau$ controls matching sharpness and each row of $\mathbf{A}_t$ specifies a correspondence distribution over the reference template.
Rather than learning the assignment independently, we jointly predict a canonical deformation field $\Delta\mathbf{P}^{\mathrm{ref}}_t\in\mathbb{R}^{M\times 3}$ and a global rigid transformation $(\mathbf{R}_t,\boldsymbol{\tau}_t)$.
The reference point cloud is accordingly adapted to the current object geometry as
\begin{equation}
\widehat{\mathbf{P}}^{\mathrm{ref}}_t
=
(\mathbf{P}^{\mathrm{ref}}+\Delta\mathbf{P}^{\mathrm{ref}}_t)\mathbf{R}_t^{\top}+\boldsymbol{\tau}_t.
\end{equation}
The soft assignment then associates each observed point with the adapted reference geometry:
\begin{equation}
\mathbf{X}_t
=
\mathbf{A}_t\widehat{\mathbf{P}}^{\mathrm{ref}}_t.
\end{equation}
By coupling point assignment with the predicted deformation and rigid transformation, correspondences are structurally constrained by underlying geometry rather than learned from feature similarity alone.
This geometric consistency underpins the correspondence losses introduced next.

\textbf{Correspondence Learning.}
To promote motion- and deformation-invariant point correspondences, we train the correspondence network on paired RGB-D observations $(s,t)$ using shared weights.
For each pair, pseudo-ground-truth pixel correspondences $\mathcal{M}_{st}=\{(i,j)\}$ are generated via MARCO~\cite{marco}. 
Specifically, we map each observation point to the undeformed canonical template as $\mathbf{Z}_r = \mathbf{A}_r\mathbf{P}^{\mathrm{ref}}\in\mathbb{R}^{N_r\times3}, r\in\{s,t\}$,
and minimize
\begin{equation}
    \mathcal{L}_{\mathrm{pair}}
    =
    \frac{1}{|\mathcal{M}_{st}|}
    \sum_{(i,j)\in\mathcal{M}_{st}}
    \rho\left(\mathbf{Z}_s^i, \mathbf{Z}_t^j\right),
\end{equation}
where $\rho(\cdot)$ denotes the SmoothL1 loss~\cite{fast-rcnn}. 
This objective encourages matched pixels to align with the same canonical template region, regardless of their current 3D positions.

For each observation $r$, we further impose a geometric consistency objective
\begin{equation}
    \mathcal{L}_{\mathrm{geo}}
    =
    \frac{\lambda_{\mathrm{cd}}}{2}
    \sum_{r\in\{s,t\}}
    \mathcal{L}_{\mathrm{cd}}
    (
        \mathbf{X}_r,
        \widehat{\mathbf{P}}^{\mathrm{ref}}_r
    )
    +
    \frac{\lambda_{\mathrm{3D}}}{2}
    \sum_{r\in\{s,t\}}
    \rho\left(\mathbf{X}_r, \mathbf{P}_r\right),
\end{equation}
where $\mathcal{L}_{\mathrm{cd}}$ represents bidirectional Chamfer distance~\cite{cdloss} that enforces template coverage, while the second term aligns template-mediated coordinates with observed 3D points.

Projection supervision is structured as
\begin{equation}
    \mathcal{L}_{\mathrm{proj}}
    =
    \lambda_{\mathrm{kp}}\mathcal{L}_{\mathrm{kp}}
    +\lambda_{\mathrm{rep}}\mathcal{L}_{\mathrm{rep}}
    +\lambda_{\mathrm{mask}}\mathcal{L}_{\mathrm{mask}}
    +\lambda_{\mathrm{fill}}\mathcal{L}_{\mathrm{fill}}.
\end{equation}
Here, $\mathcal{L}_{\mathrm{kp}}$ aligns projected template keypoints with annotated image keypoints, and $\mathcal{L}_{\mathrm{rep}}$ re-projects $\mathcal{X}_r$ back to its corresponding image pixels. 
The silhouette terms utilize a mask distance field to constrain the projected template within the object mask ($\mathcal{L}_{\mathrm{mask}}$) and an occupancy objective to encourage full foreground coverage ($\mathcal{L}_{\mathrm{fill}}$).

Finally, $\mathcal{L}_{\mathrm{reg}}$ penalizes excessive deformation, preserves local geometry through an ARAP-inspired regularizer~\cite{arap}, and discourages assignment collapse. 

Altogether, our complete training objective is defined as
\begin{equation}
\mathcal{L}
=
\lambda_{\mathrm{pair}}\mathcal{L}_{\mathrm{pair}}
+\lambda_{\mathrm{geo}}\mathcal{L}_{\mathrm{geo}}
+\lambda_{\mathrm{proj}}\mathcal{L}_{\mathrm{proj}}
+\lambda_{\mathrm{reg}}\mathcal{L}_{\mathrm{reg}}.
\end{equation}

\textbf{Grasp Transfer and Reconstruction.}
For a given initial grasp $\mathbf{T}_s^g$ at time $s$, we recover the corresponding grasp $\mathbf{T}_t^g$ at time $t$ under scene variations.
We represent the grasp by two intended contact points on the object and transfer them through the shared template correspondence space.
Leveraging soft assignments
$\mathbf{A}_s\in\mathbb{R}^{N_s\times M}$ and
$\mathbf{A}_t\in\mathbb{R}^{N_t\times M}$, we construct the
source-to-target correspondence affinity as
\begin{equation}
\mathbf{C}_{s\rightarrow t}
=
\mathbf{A}_s\mathbf{A}_t^\top
\in\mathbb{R}^{N_s\times N_t},
\end{equation}
where $\mathbf{C}_{s\rightarrow t}(i,j)=\left\langle
        \mathbf{A}_s(i,:),
        \mathbf{A}_t(j,:)
    \right\rangle$
measures the similarity between source point $i$ and target point $j$.
The shared template enables this construction, while Sinkhorn normalization encourages balanced marginals and reduces many-to-one assignment collapse.
Let $\mathbf{c}_{s,1},\mathbf{c}_{s,2}\in\mathbb{R}^{3}$ denote the source grasp contacts. 
Each contact is matched to the highest-scoring target point in the corresponding row of $\mathbf{C}_{s\rightarrow t}$, yielding $\mathbf{c}_{t,1}$ and $\mathbf{c}_{t,2}$.

To reconstruct the grasp pose, the intermediate grasp center $\tilde{\mathbf{p}}_t^g$ and closing direction $\mathbf{d}_t^g$ are computed as:
\begin{equation}
\tilde{\mathbf{p}}_t^g
=
\frac{\mathbf{c}_{t,1}+\mathbf{c}_{t,2}}{2},
\qquad
\mathbf{d}_t^g
=
\frac{\mathbf{c}_{t,2}-\mathbf{c}_{t,1}}
{\|\mathbf{c}_{t,2}-\mathbf{c}_{t,1}\|_2},
\end{equation}
where a height correction is applied to $\tilde{\mathbf{p}}_t^g$ to obtain the final center $\mathbf{p}_t^g$.
Aligning the remaining orientations with the initial grasp constructs $\mathbf{R}_t^g\in\mathrm{SO}(3)$, forming the transferred grasp pose:
\begin{equation}
\mathbf{T}_t^g
=
\begin{bmatrix}
\mathbf{R}_t^g & \mathbf{p}_t^g\\
\mathbf{0}^{\top} & 1
\end{bmatrix}
\in \mathrm{SE}(3).
\end{equation}
The reconstructed grasp and its confidence are then passed to the Information Interaction Module for feasibility checking and subsequent execution or fallback routing.

\subsection{Information Interaction Module}
The \emph{Information Interaction Module} bridges semantic grounding, geometric adaptation and execution.
It converts semantic outputs into initial grasps, validates geometrically adapted poses, and routes failures back for replanning.

\textbf{Grasp Initialization.}
Using the object point cloud and the interaction-grounded local mask, we employ AnyGrasp~\cite{anygrasp} to generate candidate grasps.
Candidates with contacts outside the mask are discarded, and the top-scoring remaining pose becomes the initial $\mathbf{T}_s^g$.
Its two surface contacts $\mathbf{c}_{s,1}$ and $\mathbf{c}_{s,2}$ are retained as task-relevant anchors for subsequent geometric adaptation, as detailed in Section~\ref{sec:geometric_adaptation}.

\begin{figure}[t]
    \centering
    \includegraphics[width=\columnwidth]{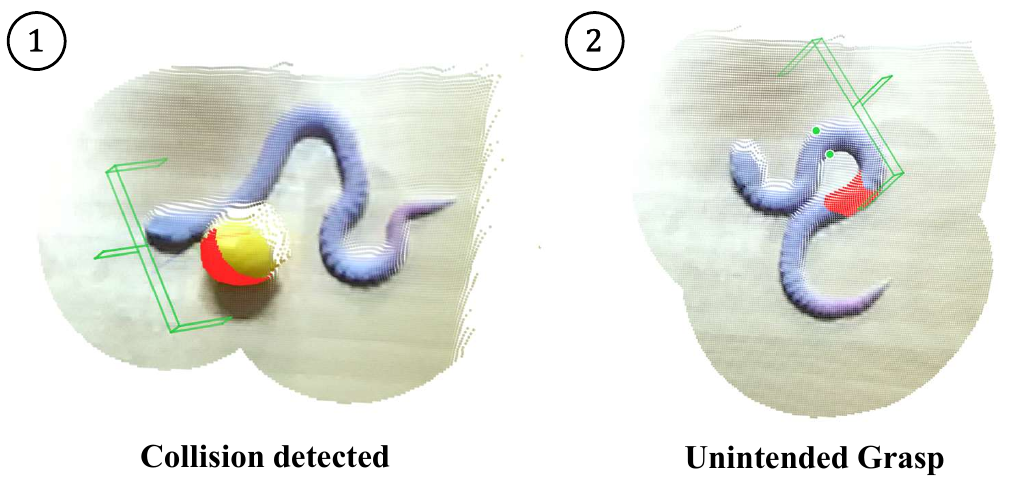}
    \caption{
        Illustration of feasibility checking.
        The green gripper denotes the updated grasp, green points the matched grasp contacts, and red markers infeasible interactions. 
        (1) Collision-prone and (2) unintended grasps are rejected before execution, triggering semantic replanning.
    }
    \label{fig:routing}
    \vspace{-4mm}
\end{figure}

\textbf{Feasibility Checking.}
Each reconstructed grasp is evaluated by its correspondence confidence, reachability, and collision safety.
Reachability checking verifies whether the transferred pose is kinematically feasible, while collision checking detects collisions based on the current scene point cloud during gripper approach and closing, including interference with surroundings or unintended parts of the target object.
Fig.~\ref{fig:routing} illustrates two cases.
A feasible grasp is directly executed, whereas unreliable correspondence or failed checks trigger fallback to the semantic path for replanning.

\section{EXPERIMENTS}
\label{sec:experiments}

\subsection{Experimental Setup}

\textbf{Hardware.}
All real-world experiments use a KUKA LBR iiwa 14 R820 arm with a Robotiq 2F-140 gripper. Two ZED 2i cameras provide complementary RGB-D views, one wrist-mounted and one fixed at an elevated viewpoint opposite the robot. Local perception and control modules run on a workstation with an NVIDIA GeForce RTX 3090 GPU.

\textbf{Evaluation Settings.}
We evaluate DualManip across six manipulation tasks spanning non-rigid deformation, articulated reconfiguration, rigid motion, and precision assembly, categorized into three general tasks and three PC chassis assembly tasks.
The general tasks require the robot to \emph{``put a soft snake into a bowl''} (\texttt{snake-placement}, non-rigid), \emph{``position an Ultraman toy between the monster and astronaut figurines''} (\texttt{toy-positioning}, articulated), and \emph{``place a block fish into a bucket''} (\texttt{block-placement}, rigid).
The assembly tasks include \texttt{RAM Insertion} and \texttt{GPU Insertion}, which require inserting a memory module (RAM) and a graphics card (GPU) into fixed chassis slots, respectively, and \texttt{Fan Placement}, which requires placing a cooling fan at its designated mounting location.

We consider three settings: 
(1) \emph{static}, where the scene remains unchanged throughout execution, providing a baseline measure of system capability;
(2) \emph{single-change}, where the object undergoes a single motion or deformation during execution and then remains stationary; and
(3) \emph{continuous dynamic}, where the object moves or deforms continuously during execution.
The three PC assembly tasks are tested only under the static setting, as they primarily assess precision assembly rather than dynamic adaptation, whereas the general tasks are tested under all three settings. 
Each object-condition-method combination is evaluated over 15 trials.

\textbf{Implementation Details.}
For each object, a reference template is constructed prior to evaluation and kept fixed across all trials.
Both the template and observed point clouds are downsampled to 1,024 points for correspondence estimation, i.e., $M=N_t=1024$.
To train the correspondence network, we collect 100 RGB-D observations per object from each camera viewpoint.
We set $\lambda_{\mathrm{pair}}=100$, $(\lambda_{\mathrm{cd}},\lambda_{\mathrm{3D}})=(10,100)$, $(\lambda_{\mathrm{kp}},\lambda_{\mathrm{rep}},\lambda_{\mathrm{mask}},\lambda_{\mathrm{fill}})=(10,10,60,2)$, and $\lambda_{\mathrm{geo}}=\lambda_{\mathrm{proj}}=\lambda_{\mathrm{reg}}=1$.
GPT-5.6 Terra~\cite{gpt5-6} is accessed via the OpenAI API and used as the VLM for semantic reasoning in all experiments, while Qwen3.5-2B~\cite{qwen3.5} is additionally deployed locally for agentic latency evaluation.

\begin{table*}[t]
\centering
\caption{Quantitative results under the static setting.
Entries denote successful trials, while \emph{Mean} reports average success rate (\%).}
\label{tab:static_results}

\renewcommand{\arraystretch}{1.00}
\setlength{\aboverulesep}{0.3ex}
\setlength{\belowrulesep}{0.3ex}
\begin{tabular*}{0.95\textwidth}{
@{\extracolsep{\fill}}
llcccc
@{\hspace{2pt}}
}
\toprule
\textbf{Category} &
\textbf{Task} &
\textbf{ReKep} &
\textbf{OmniManip} &
\textbf{CLEA} &
\textbf{DualManip (Ours)} \\
\midrule

\multirow[t]{4}{*}{General}
& Snake Placement
& 8/15  & \textbf{12/15} & 11/15 & \textbf{12/15} \\

& Toy Positioning
& 10/15 & 11/15 & \textbf{12/15} & \textbf{12/15} \\

& Block Placement
& 9/15 & \textbf{12/15} & 11/15 & 11/15 \\

& \textit{Mean}
& 60.0\% & \textbf{77.8\%} & 75.6\% & \textbf{77.8\%} \\

\midrule

\multirow[t]{4}{*}{Assembly}
& RAM Insertion
& 1/15 & 0/15 & 0/15 & \textbf{4/15} \\

& GPU Insertion
& 4/15 & 6/15 & 5/15 & \textbf{7/15} \\

& Fan Placement
& 9/15 & \textbf{12/15} & 11/15 & \textbf{12/15} \\

& \textit{Mean}
& 31.1\% & 40.0\% & 35.6\% & \textbf{51.1\%} \\

\bottomrule
\end{tabular*}
\vspace{-4mm}
\end{table*}

\subsection{Baselines}
We compare DualManip with four baselines representing distinct responses to scene changes.
(1) \textbf{ReKep}~\cite{rekep} performs closed-loop execution through sparse keypoint tracking~\cite{cotracker3}.
(2) \textbf{OmniManip}~\cite{omnimanip} represents rigid 6D pose-based adaptation, using object-level pose tracking to update the manipulation target during execution.
(3) \textbf{CLEA}~\cite{clea} represents agentic VLM-based adaptation, addressing scene changes through visual validation and high-level replanning rather than direct geometric adaptation.
(4) \textbf{Open-loop} executes the initial grasp without geometric adaptation, directly evaluating the impact of omitting scene changes. It is evaluated only under dynamic settings, as no adaptation is needed in static scenes.
All methods share the same execution backend where applicable to isolate their scene-adaptation mechanisms.

\begin{figure}[t]
    \centering
    \includegraphics[width=\columnwidth]{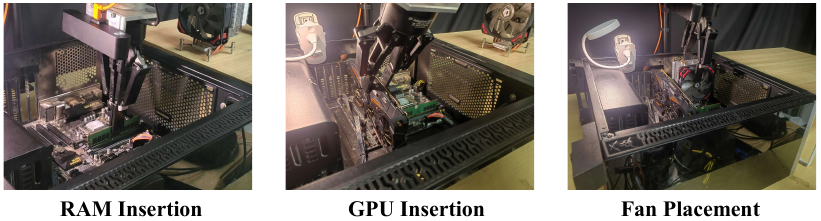}
    \caption{
        Representative execution results for the high-precision assembly tasks under the static setting, highlighting DualManip's ability to perform precise task-conditioned insertion and placement.
    }
    \label{fig:assembly}
    \vspace{-4mm}
\end{figure}

\subsection{Static Manipulation Results}
We first evaluate all methods under the static setting to assess their baseline manipulation capability.
Table~\ref{tab:static_results} shows that DualManip matches the best baseline on general tasks (77.8\%) and achieves the highest mean success rate on assembly tasks (51.1\%, versus 40.0\% for the strongest baseline). 
Fig.~\ref{fig:assembly} presents the execution results of the assembly tasks.
Overall, these results verify that DualManip maintains competitive static manipulation performance  before we turn to its primary capability of dynamic adaptation.

\begin{table*}[t]
\centering
\caption{Quantitative results under dynamic settings.
Entries denote successful trials, while \emph{Mean} reports average success rate (\%).}
\label{tab:dynamic_results}

\renewcommand{\arraystretch}{1.00}
\setlength{\aboverulesep}{0.3ex}
\setlength{\belowrulesep}{0.3ex}

\begin{tabular*}{0.95\textwidth}{
@{\extracolsep{\fill}}
llccccc
@{\hspace{2pt}}
}
\toprule
\textbf{Setting} &
\textbf{Task} &
\textbf{ReKep} &
\textbf{OmniManip} &
\textbf{CLEA} &
\textbf{Open-loop} &
\textbf{DualManip (Ours)} \\
\midrule

\multirow[t]{4}{*}{Single-change}
& Snake Placement
& 5/15 & 2/15 & 9/15 & 0/15 & \textbf{10/15} \\

& Toy Positioning
& 7/15 & 1/15 & \textbf{10/15} & 0/15 & 9/15 \\

& Block Placement
& 7/15 & 9/15 & \textbf{10/15} & 0/15 & \textbf{10/15} \\

& \textit{Mean}
& 42.2\% & 26.7\% & \textbf{64.4\%} & 0.0\% & \textbf{64.4\%} \\

\midrule

\multirow[t]{4}{*}{Continuous}
& Snake Placement
& 3/15 & 1/15 & 0/15 & 0/15 & \textbf{8/15} \\

& Toy Positioning
& 4/15 & 0/15 & 0/15 & 0/15 & \textbf{7/15} \\

& Block Placement
& 6/15 & 7/15 & 0/15 & 0/15 & \textbf{9/15} \\

& \textit{Mean}
& 28.9\% & 17.8\% & 0.0\% & 0.0\% & \textbf{53.3\%} \\

\bottomrule
\end{tabular*}
\vspace{-4mm}
\end{table*}

\subsection{Dynamic Manipulation Results}
\begin{figure}[t]
    \centering
    \includegraphics[width=\columnwidth]{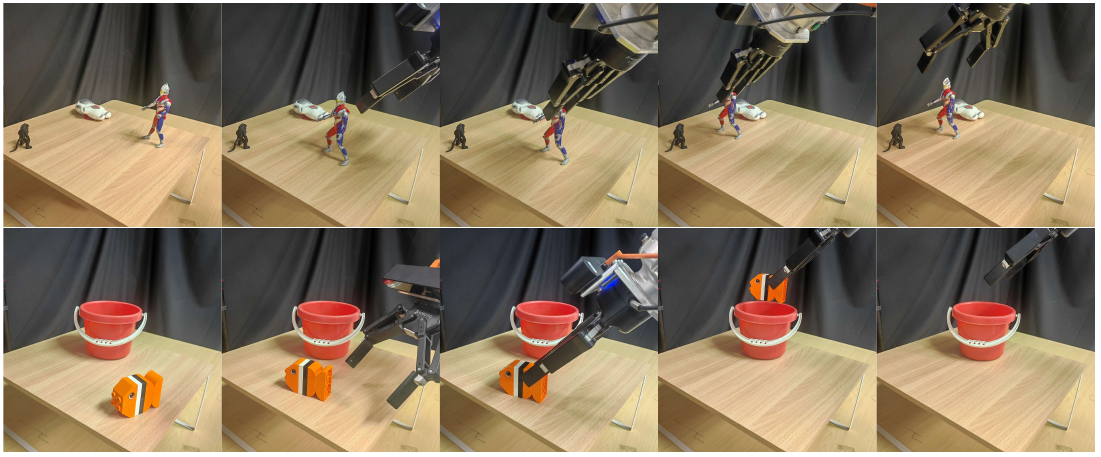}

    \caption{
        Representative executions under the single-change setting.
        Top: \texttt{toy-positioning} under articulated reconfiguration.
        Bottom: \texttt{block-placement} under rigid displacement.
    }
    \label{fig:single_change_results}
    \vspace{-4mm}
\end{figure}

We next evaluate adaptation under single-change and continuous dynamic scene variations, as summarized in Table~\ref{tab:dynamic_results}.
Under the single-change setting, DualManip and CLEA both achieve the highest mean success rate of 64.4\%, outperforming ReKep (42.2\%) and OmniManip (26.7\%).
OmniManip degrades drastically on the articulated toy and deformable snake, exposing the limitation of object-level rigid pose tracking under articulated and non-rigid changes.
ReKep better handles non-rigid motion through sparse keypoint tracking, but remains sensitive to unreliable tracks on weakly textured or strongly deforming regions.
In contrast, DualManip directly transfers task-relevant contacts through a shared template space, enabling effective and precise adaptation to diverse geometric changes.
As shown in Fig.~\ref{fig:single_change_results}, the transferred grasp remains consistent with the original task intent after articulated reconfiguration or rigid displacement.
The comparable performance of CLEA indicates that high-level replanning can also recover from an isolated scene change, but at considerably higher computational cost, as discussed later.

\begin{figure}[t]
    \centering
    \includegraphics[width=\columnwidth]{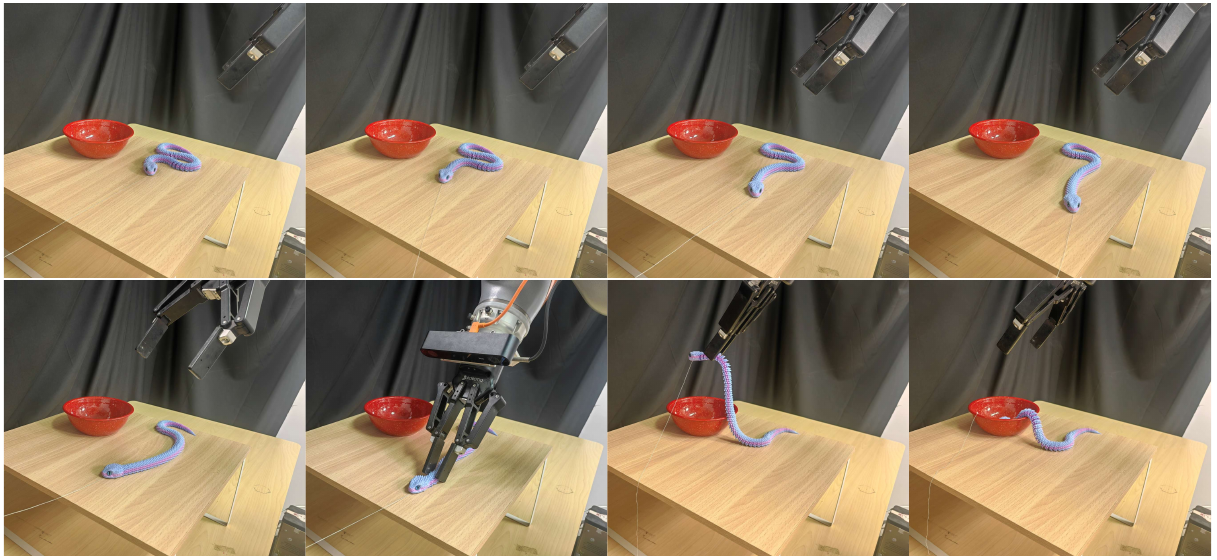}
    \caption{
        Qualitative execution on \texttt{snake-placement} under continuous scene variations. DualManip continuously updates grasp poses in real time to accommodate object motion and deformation while maintaining valid task intent.
    }
    \label{fig:continuous_results}
    \vspace{-4mm}
\end{figure}

The advantage of DualManip becomes more pronounced under continuous scene changes.
DualManip maintains a 53.3\% mean success rate, considerably surpassing ReKep (28.9\%) and OmniManip (17.8\%), whereas CLEA and Open-loop fail completely.
Continuous scene changes further amplify the limitations of sparse point tracking and rigid pose representations, as small geometric mismatches can accumulate throughout execution.
CLEA, despite its robust replanning capability, cannot keep pace with continuously evolving geometry due to the latency of repeated high-level reasoning.
In contrast, DualManip continuously updates task-relevant contacts via its geometric path, seamlessly adapting execution to evolving geometry while bypassing repetitive VLM inference.
Figure~\ref{fig:continuous_results} showcases this behavior on the deformable snake, where the grasp is continuously adjusted as the object moves and deforms.

\begin{table}[t]
\centering
\caption{Quantitative comparison of average adaptation latency under a single scene change.}
\label{tab:efficiency}
\renewcommand{\arraystretch}{1.08}
\setlength{\tabcolsep}{3.9pt} 

\begin{tabular}{@{}lccccc@{}}
\toprule
\multirow{2}{*}{\textbf{Method}} &
\multirow{2}{*}{\textbf{ReKep}} &
\multirow{2}{*}{\textbf{OmniManip}} &
\multicolumn{2}{c}{\textbf{CLEA}} &
\multirow{2}{*}{\textbf{DualManip (Ours)}} \\
\cmidrule(lr){4-5}

&
&
&
\textbf{GPT} &
\textbf{Qwen} &
\\
\midrule

\textbf{Latency}
& 77.2\,ms
& 231.3\,ms
& 6.376\,s
& 4.329\,s
& 138.7\,ms \\

\bottomrule
\end{tabular}
\vspace{-4mm}
\end{table}

\subsection{Latency Analysis}
A central goal of DualManip is to handle frequent geometric changes without repeated high-latency semantic reasoning.
We therefore compare adaptation latency, measured from scene update arrival to internal update completion.
For DualManip, this includes correspondence estimation and grasp reconstruction, whereas ReKep measures keypoint tracking.
Excluding shared downstream modules with large runtime variance, we report the average latency over 20 independent updates for each method.

Table~\ref{tab:efficiency} reports the average adaptation latency of the compared methods under a single scene change.
DualManip requires 138.7\,ms per adaptation, comparable in latency to ReKep and OmniManip while achieving stronger performance under scene variations.
In contrast, CLEA requires 6.376\,s for state verification and semantic replanning with GPT~\cite{gpt5-6}.
We additionally evaluate CLEA with a locally deployed Qwen3.5-2B~\cite{qwen3.5} model to assess adaptation latency without remote API calls, yet it still has an adaptation latency of 4.329\,s, approximately $31\times$ that of DualManip.
These results highlight DualManip's strength in combining low-latency adaptation with robust dynamic manipulation.


\section{CONCLUSION}
We presented DualManip, a dual-path manipulation framework that decouples high-level semantic reasoning from low-latency geometric adaptation under scene dynamics. 
By validating shape-adaptive contact transfers prior to execution, DualManip maintains task intent while efficiently adapting to rigid motion, articulated reconfiguration, and non-rigid deformation. 
Extensive real-world experiments confirm its superior manipulation robustness and markedly lower adaptation latency compared to agentic verification and replanning.

\textbf{Limitations.} 
DualManip relies on pre-constructed templates and object-specific training for dense correspondence, constraining direct zero-shot generalization to unseen instances. 
Future work will explore category-level correspondence learning to relax template dependencies while preserving real-time responsiveness.


\bibliographystyle{IEEEtran}
\bibliography{reference}

\end{document}